\documentclass{article}
\usepackage{spconf,amsmath,graphicx,hyperref,amsfonts,enumitem}
\usepackage{soul,xcolor}

\usepackage{colortbl, multirow, bbding, arydshln, booktabs}
\usepackage[flushleft]{threeparttable}
\newcommand\Greentext[1]{\textcolor[RGB]{93, 141, 74}{#1}}
\usepackage{pifont}

\title{TinyDrop: Tiny Model Guided Token Dropping for Vision Transformers}
\name{Guoxin Wang$^{\star}$ \qquad Qingyuan Wang$^{\star}$ \qquad Binhua Huang$^{\star}$ \qquad Shaowu Chen$^{\dagger}$ \qquad Deepu John$^{\star}$}
  
\address{$^{\star}$School of Electrical and Electronic Engineering, University College Dublin, Ireland\\
$^{\dagger}$College of Electronics and Information Engineering, Shenzhen University, China}
\begin{document}
%
\maketitle
\begin{abstract}
Vision Transformers (ViTs) achieve strong performance in image classification but incur high computational costs from processing all image tokens. To reduce inference costs in large ViTs without compromising accuracy, we propose TinyDrop, a training-free token dropping framework guided by a lightweight vision model. The guidance model estimates the importance of tokens while performing inference, thereby selectively discarding low-importance tokens if large vit models need to perform attention calculations. The framework operates plug-and-play, requires no architectural modifications, and is compatible with diverse ViT architectures. Evaluations on standard image classification benchmarks demonstrate that our framework reduces FLOPs by up to 80\% for ViTs with minimal accuracy degradation, highlighting its generalization capability and practical utility for efficient ViT-based classification.
\end{abstract}
\section{Introduction}
Vision Transformers (ViTs) achieve competitive image classification performance by modeling long-range dependencies among image patches through self-attention \cite{dosovitskiy2021an, liu2022swin}. Unlike convolutional networks, ViTs process all input tokens uniformly with identical computational cost, resulting in quadratic complexity relative to token count. While enabling effective global context capture, this approach introduces computational redundancy as many tokens correspond to low-information regions (e.g., uniform backgrounds). For large ViT models such as ViT$_\mathit{L/16}$ \cite{dosovitskiy2021an}, BEiTv2$_\mathit{L}$ \cite{peng2022beit} and DeiT3$_\mathit{L}$ \cite{touvron2022deit}, this redundancy presents significant deployment bottlenecks in resource-constrained or real-time applications.

Recent work reduces ViT inference cost via knowledge distillation, low-rank/structured approximations, and token compression \cite{sun2024logit, song2024low, bolya2023token}. Among these, token-level methods are particularly effective, which prunes or merges tokens before self-attention. Training-based approaches learn layer-wise rates but require fine-tuning, which is undesirable for frozen backbones \cite{chen2023diffrate, fayyaz2022adaptive}. 
Training-free token reduction avoids retraining, but existing methods still hinge on internal heuristics and globally fixed budgets. ToMe \cite{bolya2023token} merges pairs under a preset per-layer schedule. While tokens get merged is data-driven, the amount and layer-wise schedule are input-agnostic, limiting per-sample budget control. Zero-TP \cite{wang2024zero} scores tokens via attention-graph procedures and prunes with a global keep ratio; the pipeline introduces extra hyperparameter choices and can be brittle across heads/backbones. PaPr \cite{mahmud2024papr} predicts one-shot patch importance with a lightweight ConvNet, yet applies a fixed keep rate, and the ConvNet's complexity was not counted. In addition, Papr reports primarily on smaller/earlier backbones, with limited accounting of cost when scaling to large frozen ViTs.


We propose TinyDrop, a training-free token dropping framework guided by a lightweight vision model \cite{li2023rethinking, tan2021efficientnetv2}. Our approach computes token importance scores using attention maps or class activation maps \cite{chen2023extracting} from the guidance model, which are resampled to align with the target ViT's token grid. The least important tokens are discarded prior to self-attention computation in the target ViT, proportionally reducing FLOPs without architectural modifications. In addition, we applied an early exit paradigm when guiding model inference \cite{wang2024tiny}, i.e., determining whether to perform target vit inference, which further reduced the overall FLOPs of the framework. This framework supports diverse ViT backbones without retraining. Comprehensive evaluations on standard benchmarks demonstrate up to 80\% FLOPs reduction with minimal accuracy degradation across ViT architectures.

The main contributions of this work are: 1) Tiny model guided token selection: Training-free framework leveraging lightweight models to exit inference and improve token importance estimation in large ViTs. 2) Plug-and-play compatibility: Model-agnostic module requiring no architectural changes or retraining of target ViTs. 3) Effective efficiency-accuracy balance: Validation showing up to 80\% FLOPs reduction with negligible accuracy loss on standard benchmarks.

\begin{figure}[!ht]
\centering
\includegraphics[width=\linewidth]{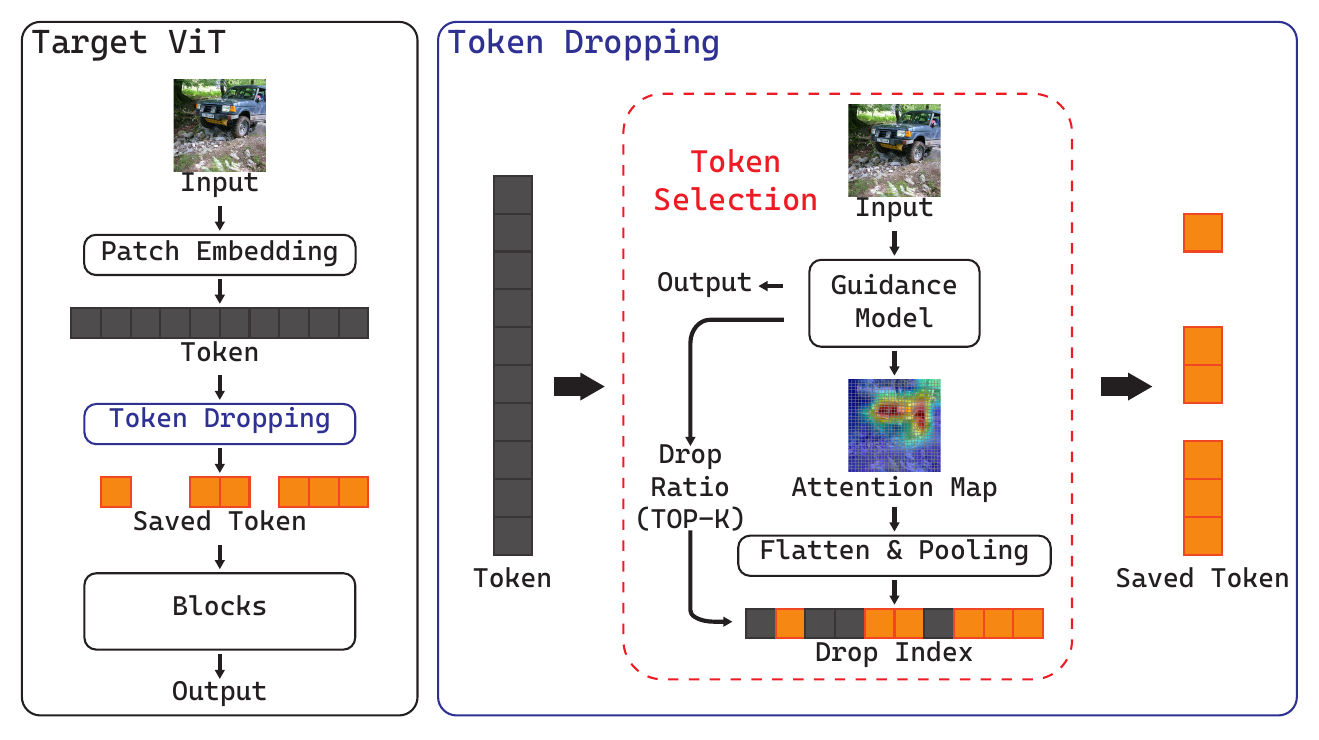}
\caption{Overview of the proposed training-free tiny model guided token dropping framework. A lightweight guidance model analyzes the input, optionally exits, and generates a saliency map, which is resized to the ViT token grid for top-K token selection. The retained tokens are then fed into the target ViT for classification with reduced computation.}
\label{fig_overview}
\end{figure}

\section{Method}
As shown in \textit{Fig. \ref{fig_overview}}, our framework operates as follows: 1) A computationally efficient guidance model $\mathcal{M}_g$ processes the input image and decides if to exit. 2) Token importance scores are derived from $\mathcal{M}_g$'s attention mechanisms. 3) Low-importance tokens are discarded before the first transformer block of the target ViT $\mathcal{M}_t$. 4) The target ViT processes only selected tokens, reducing sequence length for subsequent operations.

\subsection{Guidance Model Inference}
\subsubsection{Confidence-based Early Exit}
Given an input image, the guidance model $\mathcal{M}_g$ produces a class-confidence $c = \max(p)$, where $p$ is the softmax probability over classes. If the confidence exceeds a predefined threshold $\tau \in (0,1)$, we directly adopt the output of the guidance model as the final prediction and terminate computation; otherwise, the sample is deferred to the subsequent stages for further processing by the target model.

\subsubsection{Attention-Based Saliency Mapping}
We compute token-wise saliency maps for the guidance model $\mathcal{M}_g$ by applying Grad-CAM \cite{selvaraju2017grad} to the feature maps of its final transformer block when target ViT inference is needed. The saliency map $\mathcal{S}_g$ is obtained through the following steps. First, gradients of the model output with respect to the final block’s feature maps are calculated. These gradients are used to weight the corresponding activation channels, followed by channel-wise pooling to obtain a single saliency representation. The pooled activation map is then resampled via bilinear interpolation to a spatial resolution of $\sqrt{T}\times\sqrt{T}$, where $T$ denotes the number of patch tokens excluding the class token. Finally, we apply min–max normalization to constrain saliency values within $[0,1]$, yielding the normalized saliency map $\mathcal{S}_g \in [0,1]^{T}$ that assigns an importance score to each token.  

\subsubsection{Adaptive Confidence-to-Drop Mapping}
Token dropping is governed by the classification confidence of the guidance model as well. To adaptively determine the proportion of tokens to discard, we introduce a parametric confidence-to-drop mapping function:  

\begin{equation}
r = \min(r_{\max},r_{\max} \cdot (\dfrac{c}{\tau})^{\gamma})
\end{equation}

Here, we use the same value $\tau$ to project confidence from $[0,\tau]$ to $[0,1]$, $\gamma$ controls the curvature of the mapping function (i.e., how aggressively the drop ratio increases with confidence), and $r_{\max}$ denotes the maximum allowable drop ratio. The preserved token count for a given sample is then defined as  
\begin{equation}
K = \max \left(1, \left\lfloor (1 - r)T \right\rfloor \right),
\end{equation}
ensuring that at least one informative token is always retained.  

\subsection{Token Selection and Positional Embedding}
\subsubsection{Structured Token Selection}
Given the importance scores $\mathcal{S} \in \mathbb{R}^{T}$ and the preserved token count $K$, the indices of the top-$K$ most informative tokens are determined as
\begin{equation}
\mathcal{S}_{\mathrm{keep}} = \mathrm{argTopK}(\mathcal{S}, K).
\end{equation}

Let $X_{\mathrm{patch}} \in \mathbb{R}^{B \times T \times C}$ denote the patch embeddings of the target model $\mathcal{M}_t$, where $B$ is the batch size and $C$ is the channel dimension. The embeddings corresponding to the selected tokens are extracted as
\begin{equation}
X_{\mathrm{kept}} = \{ X_{\mathrm{patch}}[i] \mid i \in \mathcal{S}_{\mathrm{keep}} \} \in \mathbb{R}^{B \times K \times C}.
\end{equation}

The class token embedding $X_{\mathrm{cls}} \in \mathbb{R}^{B \times 1 \times C}$ is then prepended, yielding the reduced token sequence
\begin{equation}
X' = \mathrm{Concat}(X_{\mathrm{cls}}, X_{\mathrm{kept}}) \in \mathbb{R}^{B \times (K+1) \times C}.
\end{equation}

\begin{table*}[!ht]
\centering
\setlength{\tabcolsep}{7pt}
\caption{Performance comparison of the proposed training-free token dropping framework with different guidance models on three large-scale ViT architectures. For each target model, the average FLOPs (including guidance model forward and Grad-CAM backward) are reported when the accuracy drop is within 0.3\%, 0.5\%, and 1\%, respectively.}
\label{tab_result}
\resizebox{\linewidth}{!}{
\begin{tabular}{lcl|lc|lll}
\hline \hline
\multicolumn{3}{l|}{Target Model} & \multicolumn{2}{l|}{Guidance Model} & \multicolumn{3}{l}{TinyDrop GFLOPs($\Delta$) when Acc Drop $\leq$ } \\
Name & \multicolumn{1}{l}{GFLOPs} & Acc & Name & \multicolumn{1}{l|}{GFLOPs} & 1\% & 0.5\% & 0.3\% \\ \hline
\multirow{2}{*}{ViT$_\mathit{L/16}$} & \multirow{2}{*}{61.6} & \multirow{2}{*}{85.8\%} & EfficientFormerV2$_\mathit{s2}$ & 1.2 & \ \ 8.0 \Greentext{(-87.0\%)} & 11.1 \Greentext{(-82.0\%)} & 14.0 \Greentext{(-77.3\%)} \\
 &  &  & EfficientNetV2$_\mathit{rw/t}$ & 1.9 & 11.4 \Greentext{(-81.5\%)} & 14.5 \Greentext{(-75.0\%)} & 21.5 \Greentext{(-65.1\%)} \\
\hdashline
\multirow{2}{*}{BEiTv2$_\mathit{L}$} & \multirow{2}{*}{44.8} & \multirow{2}{*}{88.4\%} & EfficientFormerV2$_\mathit{s2}$ & 1.2 & \ \ 9.7 \Greentext{(-78.2\%)} & 12.8 \Greentext{(-71.3\%)} & 16.5 \Greentext{(-63.2\%)} \\
 &  &  & EfficientNetV2$_\mathit{rw/t}$ & 1.9 & 13.3 \Greentext{(-70.3\%)} & 20.8 \Greentext{(-53.6\%)} & 28.5 \Greentext{(-36.4\%)} \\
\hdashline
\multirow{2}{*}{DeiT3$_\mathit{L}$} & \multirow{2}{*}{61.6} & \multirow{2}{*}{87.0\%} & EfficientFormerV2$_\mathit{s2}$ & 1.2 & 10.9 \Greentext{(-82.3\%)} & 16.0 \Greentext{(-74.0\%)} & 28.3 \Greentext{(-54.1\%)} \\
 &  &  & EfficientNetV2$_\mathit{rw/t}$ & 1.9 & 15.1 \Greentext{(-75.5\%)} & 21.0 \Greentext{(-65.9\%)} & 38.4 \Greentext{(-37.7\%)} \\ \hline \hline
\end{tabular}
}
\end{table*}

\subsubsection{Positional Embedding Adaptation}
Token dropping alters the sequence length and ordering of tokens, thus requiring adaptation of positional embeddings \cite{kazemnejad2023impact}. Let $\mathcal{P} \in \mathbb{R}^{(T+1) \times C}$ denote the original positional embedding matrix, including the class token. The updated positional embeddings are obtained by selecting entries consistent with the retained token indices:
\begin{equation}
\mathcal{P}' = \{\mathcal{P}[0,:]\} \cup \{\mathcal{P}[i+1,:] \mid i \in \mathcal{S}_{\mathrm{keep}} \} \in \mathbb{R}^{(K+1) \times C}.
\end{equation}

For architectures employing relative position bias \cite{wu2021rethinking} $\mathcal{B} \in \mathbb{R}^{(T+1) \times (T+1) \times n_H}$, where $n_H$ is the number of attention heads, the corresponding sub-matrix is selected as
\begin{equation}
\mathcal{B}' = \mathcal{B}[\mathcal{S}_{\mathrm{keep}}, \mathcal{S}_{\mathrm{keep}}, :] \in \mathbb{R}^{(K+1) \times (K+1) \times n_H}.
\end{equation}

This procedure ensures that both absolute positional embeddings and relative position biases remain consistent with the reduced token set, allowing the target ViT to operate seamlessly on the shortened sequence.

\section{Experiments}
\subsection{Settings}
We evaluate TinyDrop across ImageNet-1K \cite{russakovsky2015imagenet} on state-of-the-art architectures, including Vision Transformer \cite{dosovitskiy2021an, steiner2022how}, BEiTv2 \cite{peng2022beit}, DeiT3 \cite{touvron2022deit} and DeiT \cite{touvron2021training}. We also evaluate the performance with different guidance models, including EfficientNetV2 \cite{tan2021efficientnetv2} and EfficientFormerV2 \cite{li2023rethinking}. All used models are pre-trained from \cite{rw2019timm}. ImageNet-1K contains 50k validation images across 1,000 categories, all resized to $224 \times 224$ and normalized using the standard ImageNet mean and standard deviation. We sweep the drop ratio $r$ using the mapping function described above, with $\gamma=0.5$, $r_{\max}=0.7$, and varying confidence thresholds $\tau$.

\subsection{Results}
\subsubsection{Performance on Various Vision Transformers}
\textit{Table \ref{tab_result}} reports the performance of TinyDrop on three large-scale ViT architectures with different guidance models. Across all settings, TinyDrop achieves substantial reductions in FLOPs while maintaining accuracy loss within 1\%, 0.5\%, and even 0.3\%. For example, on ViT$_\mathit{L/16}$, TinyDrop guided by EfficientFormerV2$_\mathit{s2}$ reduces the computational cost from 61.6 GFLOPs to 8.0 GFLOPs (-87.0\%) with at most 1\% accuracy drop, and still achieves 14.0 GFLOPs (-77.3\%) when the accuracy drop is constrained within 0.3\%. On BEiTv2$_\mathit{L}$, FLOPs are reduced to 9.7 GFLOPs (-78.2\%) under the 1\% tolerance, and to 16.5 GFLOPs (-63.2\%) when requiring $\leq$0.3\% drop. Similar trends are observed on DeiT3$_\mathit{L}$, where FLOPs are cut by up to 82.3\% with $\leq$1\% accuracy degradation. These results demonstrate that TinyDrop consistently provides 70-87\% computational savings across diverse ViT backbones, with negligible accuracy loss, validating its generalization capability and practical efficiency.

\subsubsection{State-of-The-Art Methods Comparison}
\textit{Table \ref{tab_comparison}} shows a comparison with representative token dropping methods. Across all backbones, TinyDrop consistently achieves the largest FLOPs reduction while maintaining accuracy on par with or better than existing approaches. In particular, TinyDrop reduces computation by up to 85\% on DeiT and ViT variants, whereas prior methods typically achieve only 35-50\% savings under similar accuracy levels. Importantly, TinyDrop operates in a training-free manner and remains effective across both lightweight and large-scale ViTs. Ablation studies further show that both the token dropping and the guidance-based exit contribute to efficiency, with their combination yielding the strongest performance. These results highlight TinyDrop as a highly practical and generalizable framework for efficient ViT inference.
\definecolor{lightgray}{HTML}{888888}
\begin{table}[!ht]
\centering
\setlength{\tabcolsep}{2pt}
\caption{Comparison of TinyDrop with existing token dropping methods on ImageNet across ViT backbones. TinyDrop achieves the largest FLOPs reduction (up to 83-85\%) while preserving accuracy and requiring no training. Ablation results without early exit and without token dropping are also included.}
\label{tab_comparison}
\resizebox{\linewidth}{!}{
\begin{threeparttable}[b]
\begin{tabular}{@{}llcccc@{}}
\toprule \toprule
Model & Method & \multicolumn{1}{l}{\begin{tabular}[c]{@{}l@{}}Training\\ Free\end{tabular}} & Acc & GFLOPs & $\Delta$FLOPs \\ \midrule
& \textcolor{lightgray}{Baseline} & \textcolor{lightgray}{-} & \textcolor{lightgray}{79.8\%} & \textcolor{lightgray}{4.6} & \textcolor{lightgray}{-} \\
 & DiffRate$^\star$ \cite{chen2023diffrate} & \ding{55} & 79.6\% & 2.9 & -50.0\% \\
 & ATS \cite{fayyaz2022adaptive} & \ding{55} & \textbf{79.7\%} & 2.9 & -37.0\% \\
 & ToMe \cite{bolya2023token} & \ding{51} & 79.4\% & 2.7 & -41.3\% \\
 & ToFu \cite{kim2024token} & \ding{51} & 79.5\% & 2.7 & -41.3\% \\
 & Zero-TP \cite{wang2024zero} & \ding{51} & 79.1\% & 2.5 & -45.7\% \\
 & PaPr \cite{mahmud2024papr} & \ding{51} & 79.2\% & 3.0$^\dagger$ & -34.9\% \\
 & \cellcolor[HTML]{EFEFEF}TinyDrop & \cellcolor[HTML]{EFEFEF}\ding{51} & \cellcolor[HTML]{EFEFEF}79.6\% & \cellcolor[HTML]{EFEFEF}\textbf{1.4} & \cellcolor[HTML]{EFEFEF}\textbf{-69.6\%} \\
 & \cellcolor[HTML]{EFEFEF}w/o Early Exit & \cellcolor[HTML]{EFEFEF}\ding{51} & \cellcolor[HTML]{EFEFEF}79.1\% & \cellcolor[HTML]{EFEFEF}3.0 & \cellcolor[HTML]{EFEFEF}-34.9\% \\
 \multirow{-10}{*}{DeiT$_\mathit{S}$} & \cellcolor[HTML]{EFEFEF}w/o Token Dropping & \cellcolor[HTML]{EFEFEF}\ding{51} & \cellcolor[HTML]{EFEFEF}79.6\% & \cellcolor[HTML]{EFEFEF}\textbf{1.8} & \cellcolor[HTML]{EFEFEF}\textbf{-60.9\%} \\
\hdashline
 & \textcolor{lightgray}{Baseline} & \textcolor{lightgray}{-} & \textcolor{lightgray}{81.8\%} & \textcolor{lightgray}{17.6} & \textcolor{lightgray}{-} \\
 & DiffRate$^\star$ \cite{chen2023diffrate} & \ding{55} & 81.5\% & 11.5 & -34.7\% \\
 & ToMe \cite{bolya2023token} & \ding{51} & 81.4\% & 11.5 & -34.7\% \\
 & Zero-TP \cite{wang2024zero} & \ding{51} & 81.0\% & 13.6 & -22.7\% \\
 & \cellcolor[HTML]{EFEFEF}TinyDrop & \cellcolor[HTML]{EFEFEF}\ding{51} & \cellcolor[HTML]{EFEFEF}\textbf{81.6\%} & \cellcolor[HTML]{EFEFEF}\textbf{2.9} & \cellcolor[HTML]{EFEFEF}\textbf{-83.5\%} \\
 & \cellcolor[HTML]{EFEFEF}w/o Early Exit & \cellcolor[HTML]{EFEFEF}\ding{51} & \cellcolor[HTML]{EFEFEF}\textbf{81.6\%} & \cellcolor[HTML]{EFEFEF}12.5 & \cellcolor[HTML]{EFEFEF}-29.0\% \\
 \multirow{-7}{*}{DeiT$_\mathit{B}$}& \cellcolor[HTML]{EFEFEF}w/o Token Dropping  & \cellcolor[HTML]{EFEFEF}\ding{51} & \cellcolor[HTML]{EFEFEF}\textbf{81.6\%} & \cellcolor[HTML]{EFEFEF}\textbf{4.0} & \cellcolor[HTML]{EFEFEF}\textbf{-77.3\%} \\
\hdashline
 & \textcolor{lightgray}{Baseline} & \textcolor{lightgray}{-} & \textcolor{lightgray}{84.6\%} & \textcolor{lightgray}{17.6} & \textcolor{lightgray}{-} \\
 & ToMe \cite{bolya2023token} & \ding{51} & 80.4\% & 8.8 & -50.0\% \\
 & ToFu \cite{kim2024token} & \ding{51} & 80.7\% & 8.8 & -50.0\% \\
 & PaPr \cite{mahmud2024papr} & \ding{51} & 82.1\% & 9.3$^\dagger$ & -47.2\% \\
 & \cellcolor[HTML]{EFEFEF}TinyDrop & \cellcolor[HTML]{EFEFEF}\ding{51} & \cellcolor[HTML]{EFEFEF}\textbf{83.0\%} & \cellcolor[HTML]{EFEFEF}\textbf{2.9} & \cellcolor[HTML]{EFEFEF}\textbf{-83.5\%} \\
 & \cellcolor[HTML]{EFEFEF}w/o Early Exit & \cellcolor[HTML]{EFEFEF}\ding{51} & \cellcolor[HTML]{EFEFEF}\textbf{82.2\%} & \cellcolor[HTML]{EFEFEF}\textbf{8.1} & \cellcolor[HTML]{EFEFEF}\textbf{-54.0\%} \\
 \multirow{-7}{*}{ViT$_\mathit{B/16}$} & \cellcolor[HTML]{EFEFEF}w/o Token Dropping & \cellcolor[HTML]{EFEFEF}\ding{51} & \cellcolor[HTML]{EFEFEF}\textbf{83.0\%} & \cellcolor[HTML]{EFEFEF}\textbf{3.2} & \cellcolor[HTML]{EFEFEF}\textbf{-81.8\%} \\
\hdashline
 & \textcolor{lightgray}{Baseline} & \textcolor{lightgray}{-} & \textcolor{lightgray}{85.8\%} & \textcolor{lightgray}{61.6} & \textcolor{lightgray}{-} \\
 & ToMe \cite{bolya2023token} & \ding{51} & 83.5\% & 31.0 & -49.7\% \\
 & ToFu \cite{kim2024token} & \ding{51} & 83.9\% & 31.0 & -49.7\% \\
 & PaPr \cite{mahmud2024papr} & \ding{51} & 83.9\% & 31.1$^\dagger$ & -49.5\% \\
& \cellcolor[HTML]{EFEFEF}TinyDrop & \cellcolor[HTML]{EFEFEF}\ding{51} & \cellcolor[HTML]{EFEFEF}\textbf{85.2\%} & \cellcolor[HTML]{EFEFEF}\textbf{9.6} & \cellcolor[HTML]{EFEFEF}\textbf{-84.4\%} \\ 
 & \cellcolor[HTML]{EFEFEF}w/o Early Exit & \cellcolor[HTML]{EFEFEF}\ding{51} & \cellcolor[HTML]{EFEFEF}\textbf{84.5\%} & \cellcolor[HTML]{EFEFEF}\textbf{29.4} & \cellcolor[HTML]{EFEFEF}\textbf{-58.8\%} \\
 \multirow{-7}{*}{ViT$_\mathit{L/16}$} & \cellcolor[HTML]{EFEFEF}w/o Token Dropping & \cellcolor[HTML]{EFEFEF}\ding{51} & \cellcolor[HTML]{EFEFEF}\textbf{85.2\%} & \cellcolor[HTML]{EFEFEF}\textbf{11.6} & \cellcolor[HTML]{EFEFEF}\textbf{-81.2\%} \\ \bottomrule \bottomrule
\end{tabular}
\begin{tablenotes}
\item[$^\star$] reported both w/wo fine-tuning results, we use the better one for comparison.
\item[$\dagger$] include external modules and additional step costs.
\end{tablenotes}
\end{threeparttable}
}
\end{table}

\subsubsection{Comparison with Early Exit Paradigm}
\begin{figure}[!ht]
\centering
\includegraphics[width=\linewidth]{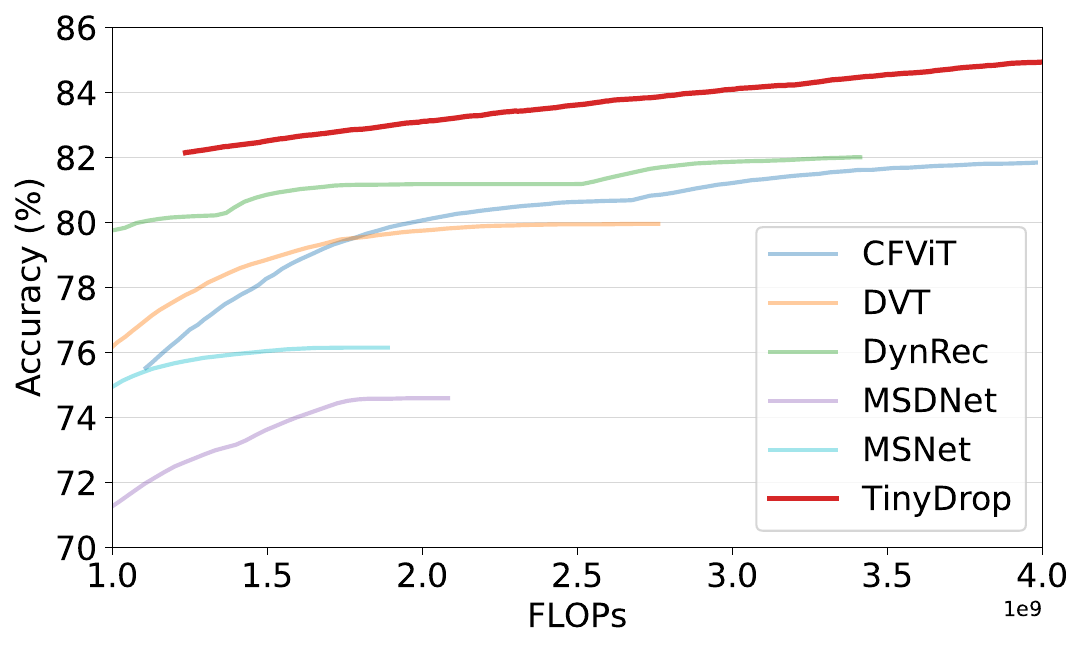}
\caption{Accuracy–FLOPs trade-off on ImageNet comparing TinyDrop with representative early exit methods. TinyDrop consistently attains higher accuracy at equal or lower FLOPs, demonstrating the efficiency of the proposed method.}
\label{fig_ee}
\end{figure}

We further compare our framework with representative early exit approaches \cite{chen2023cf, han2023dynamic, hang2022msnet, hong2021a, wang2021not}. Conventional early exit reduces computation by terminating inference in shallower layers, but typically requires architectural modifications and fine-tuning. In contrast, our method is centered on token dropping for large ViTs, which consistently achieves higher accuracy at comparable FLOPs. The lightweight guidance model also supports a simple confidence-based exit; unlike standard early exit heads, this mechanism incurs negligible overhead and is naturally integrated into our framework. As shown in \textit{Fig. \ref{fig_ee}}, TinyDrop lies consistently above existing early exit baselines, indicating that token dropping combined with lightweight guidance provides a more effective trade-off between efficiency and accuracy.

\subsection{Ablation Study}
\begin{figure}[!ht]
\centering
\includegraphics[width=\linewidth]{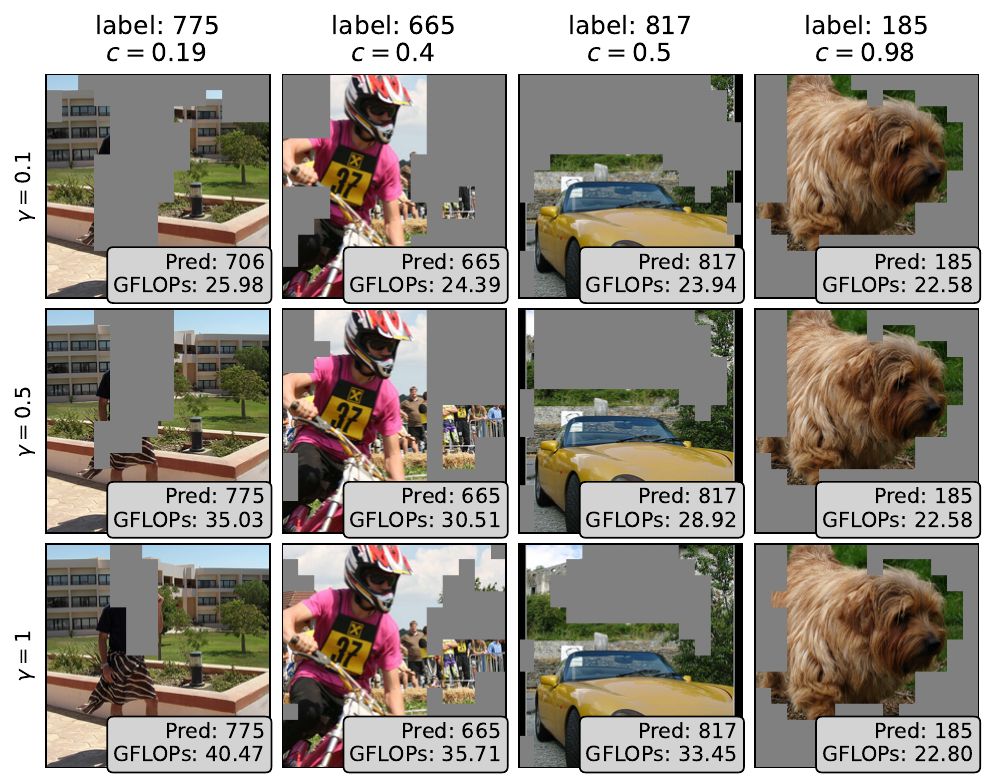}
\caption{Ablation on the curvature parameter $\gamma$ of the confidence-to-drop mapping for BEiTv2. Smaller $\gamma$ leads to more aggressive token dropping and greater FLOPs reduction but risks misclassification, while larger $\gamma$ retains more tokens and improves robustness at a higher cost. A mid-range setting achieves the best trade-off.}
\label{fig_virtu}
\end{figure}

To analyze the effect of the confidence-to-drop mapping, we conduct an ablation by varying the curvature parameter $\gamma$. \textit{Fig. \ref{fig_virtu}} visualizes representative examples across different $\gamma$ settings. A smaller $\gamma$ yields more aggressive dropping for moderately confident samples, resulting in fewer retained tokens and larger FLOPs savings. However, this may also increase the risk of discarding informative regions, leading to occasional misclassifications. Increasing $\gamma$ makes the dropping less aggressive, retaining more tokens and thereby improving robustness at the cost of higher computation. This highlights a clear trade-off: low $\gamma$ favors efficiency, high $\gamma$ favors accuracy, and a mid-range (e.g., $\gamma=0.5$) offers the best balance.

\section{Conclusion}
This paper presents a training-free token dropping framework that reduces computational costs in large Vision Transformers without retraining. By leveraging a lightweight guidance model to generate token importance scores through our confidence-to-drop mapping mechanism, we enable informed dropping while preserving the integrity of frozen pre-trained models. Extensive ImageNet evaluations demonstrate up to 80\% FLOPs reduction with $\leq1\%$ top-1 accuracy degradation across architectures. The framework is dual-agnostic—compatible with diverse ViT backbones and guidance models (both transformer- and CNN-based).

Our approach provides immediate deployment value for latency-sensitive environments due to its plug-and-play implementation and robust performance. Future work will extend this paradigm to multi-modal transformers and dense prediction tasks.

\vfill\pagebreak

\clearpage

\bibliographystyle{IEEEbib}
\bibliography{strings,refs}

\begin{thebibliography}{10}

\bibitem{dosovitskiy2021an}
Alexey Dosovitskiy et~al.,
\newblock ``An image is worth 16x16 words: Transformers for image recognition at scale,''
\newblock in {\em ICLR}, 2021.

\bibitem{liu2022swin}
Ze~Liu et~al.,
\newblock ``Swin transformer v2: Scaling up capacity and resolution,''
\newblock in {\em CVPR}, 2022, pp. 12009--12019.

\bibitem{peng2022beit}
Zhiliang Peng et~al.,
\newblock ``Beit v2: Masked image modeling with vector-quantized visual tokenizers,''
\newblock {\em arXiv preprint arXiv:2208.06366}, 2022.

\bibitem{touvron2022deit}
Hugo Touvron et~al.,
\newblock ``Deit iii: Revenge of the vit,''
\newblock in {\em ECCV}. Springer, 2022, pp. 516--533.

\bibitem{sun2024logit}
Shangquan Sun et~al.,
\newblock ``Logit standardization in knowledge distillation,''
\newblock in {\em CVPR}, 2024, pp. 15731--15740.

\bibitem{song2024low}
Lin Song et~al.,
\newblock ``Low-rank approximation for sparse attention in multi-modal llms,''
\newblock in {\em CVPR}, 2024, pp. 13763--13773.

\bibitem{bolya2023token}
Daniel Bolya et~al.,
\newblock ``Token merging: Your vit but faster,''
\newblock in {\em ICLR}, 2023.

\bibitem{chen2023diffrate}
Mengzhao Chen et~al.,
\newblock ``Diffrate: Differentiable compression rate for efficient vision transformers,''
\newblock in {\em ICCV}, 2023, pp. 17164--17174.

\bibitem{fayyaz2022adaptive}
Mohsen Fayyaz et~al.,
\newblock ``Adaptive token sampling for efficient vision transformers,''
\newblock in {\em ECCV}. Springer, 2022, pp. 396--414.

\bibitem{wang2024zero}
Hongjie Wang et~al.,
\newblock ``Zero-tprune: Zero-shot token pruning through leveraging of the attention graph in pre-trained transformers,''
\newblock in {\em CVPR}, 2024, pp. 16070--16079.

\bibitem{mahmud2024papr}
Tanvir Mahmud et~al.,
\newblock ``Papr: Training-free one-step patch pruning with lightweight convnets for faster inference,''
\newblock in {\em ECCV}. Springer, 2024, pp. 110--128.

\bibitem{li2023rethinking}
Yanyu Li et~al.,
\newblock ``Rethinking vision transformers for mobilenet size and speed,''
\newblock in {\em ICCV}, 2023, pp. 16889--16900.

\bibitem{tan2021efficientnetv2}
Mingxing Tan and Quoc Le,
\newblock ``Efficientnetv2: Smaller models and faster training,''
\newblock in {\em ICML}. PMLR, 2021, pp. 10096--10106.

\bibitem{chen2023extracting}
Zhaozheng Chen and Qianru Sun,
\newblock ``Extracting class activation maps from non-discriminative features as well,''
\newblock in {\em CVPR}, 2023, pp. 3135--3144.

\bibitem{wang2024tiny}
Qingyuan Wang et~al.,
\newblock ``Tiny models are the computational saver for large models,''
\newblock in {\em ECCV}. Springer, 2024, pp. 163--182.

\bibitem{selvaraju2017grad}
Ramprasaath~R Selvaraju et~al.,
\newblock ``Grad-cam: Visual explanations from deep networks via gradient-based localization,''
\newblock in {\em ICCV}, 2017, pp. 618--626.

\bibitem{kazemnejad2023impact}
Amirhossein Kazemnejad et~al.,
\newblock ``The impact of positional encoding on length generalization in transformers,''
\newblock {\em NeurIPS}, vol. 36, pp. 24892--24928, 2023.

\bibitem{wu2021rethinking}
Kan Wu et~al.,
\newblock ``Rethinking and improving relative position encoding for vision transformer,''
\newblock in {\em ICCV}, 2021, pp. 10033--10041.

\bibitem{russakovsky2015imagenet}
Olga Russakovsky et~al.,
\newblock ``Imagenet large scale visual recognition challenge,''
\newblock {\em International journal of computer vision}, vol. 115, no. 3, pp. 211--252, 2015.

\bibitem{steiner2022how}
Andreas~Peter Steiner et~al.,
\newblock ``How to train your vit? data, augmentation, and regularization in vision transformers,''
\newblock {\em Transactions on Machine Learning Research}, 2022.

\bibitem{touvron2021training}
Hugo Touvron et~al.,
\newblock ``Training data-efficient image transformers \& distillation through attention,''
\newblock in {\em ICML}. PMLR, 2021, pp. 10347--10357.

\bibitem{rw2019timm}
Ross Wightman,
\newblock ``Pytorch image models,'' \url{https://github.com/rwightman/pytorch-image-models}, 2019.

\bibitem{kim2024token}
Minchul Kim et~al.,
\newblock ``Token fusion: Bridging the gap between token pruning and token merging,''
\newblock in {\em WACV}, 2024, pp. 1383--1392.

\bibitem{chen2023cf}
Mengzhao Chen et~al.,
\newblock ``Cf-vit: A general coarse-to-fine method for vision transformer,''
\newblock in {\em AAAI}, 2023, pp. 7042--7052.

\bibitem{han2023dynamic}
Yizeng Han et~al.,
\newblock ``Dynamic perceiver for efficient visual recognition,''
\newblock in {\em ICCV}, 2023, pp. 5992--6002.

\bibitem{hang2022msnet}
Renlong Hang et~al.,
\newblock ``Msnet: Multi-resolution synergistic networks for adaptive inference,''
\newblock {\em IEEE Transactions on Circuits and Systems for Video Technology}, vol. 33, no. 5, pp. 2009--2018, 2022.

\bibitem{hong2021a}
Sanghyun Hong et~al.,
\newblock ``A panda? no, it's a sloth: Slowdown attacks on adaptive multi-exit neural network inference,''
\newblock in {\em ICLR}, 2021.

\bibitem{wang2021not}
Yulin Wang et~al.,
\newblock ``Not all images are worth 16x16 words: Dynamic transformers for efficient image recognition,''
\newblock {\em NeurIPS}, vol. 34, pp. 11960--11973, 2021.

\end{thebibliography}

\end{document}